\documentclass{article}
\usepackage{spconf,amsmath,graphicx, amssymb}

\usepackage{multirow} 
\usepackage{booktabs} 
\usepackage{colortbl} 
\usepackage{caption} 
\usepackage{graphicx} 

\usepackage{enumitem}
\setlist{nosep, leftmargin=14pt}

\usepackage{mwe} 

\title{Rethinking Pre-Trained Feature Extractor Selection in Multiple Instance Learning for Whole Slide Image Classification}
\name{Bryan Wong \qquad Sungrae Hong \qquad Mun Yong Yi$^{*}$\thanks{*Corresponding author: munyi@kaist.ac.kr}}
\address{Korea Advanced Institute of Science and Technology (KAIST), Daejeon, South Korea}
\begin{document}
%
\maketitle
\begin{abstract}
Multiple instance learning (MIL) has become a preferred method for gigapixel whole slide image (WSI) classification without requiring patch-level annotations. Current MIL research primarily relies on embedding-based approaches, which extract patch features using a pre-trained feature extractor and aggregate them for slide-level prediction. Despite the critical role of feature extraction, there is limited guidance on selecting optimal feature extractors to maximize WSI performance. This study addresses this gap by systematically evaluating MIL feature extractors across three dimensions: pre-training dataset, backbone model, and pre-training method. Extensive experiments were conducted on two public WSI datasets (TCGA-NSCLC and Camelyon16) using four state-of-the-art (SOTA) MIL models. Our findings reveal that: 1) selecting a robust self-supervised learning (SSL) method has a greater impact on performance than relying solely on an in-domain pre-training dataset; 2) prioritizing Transformer-based backbones with deeper architectures over CNN-based models; and 3) using larger, more diverse pre-training datasets significantly enhances classification outcomes. We hope that these insights can provide practical guidance for optimizing WSI classification and explain the reasons behind the performance advantages of the current SOTA pathology foundation models. Furthermore, this work may inform the development of more effective pathology foundation models. Our code is publicly available at \textit{https://github.com/bryanwong17/MIL-Feature-Extractor-Selection}

\end{abstract}
\section{Introduction}
\label{sec:intro}

Whole slide imaging (WSI), the process the acquiring and processing scanned images from tissue specimens for microscopic analysis, has revolutionized histopathology field by enabling remote access and high-resolution imaging. Due to their immense size, directly training deep-learning models on WSIs without preprocessing them into smaller patches is impractical with current GPUs. Moreover, annotating WSIs is time consuming and costly due to their gigapixel resolution, making it challenging for deep-learning models to predict areas of interest without available annotations.

To address these challenges, multiple instance learning (MIL) offers a weekly-supervised approach for WSI classification, which is particularly useful for pre-screening in clinical practice. MIL accommodates varying WSI sizes and requires only slide-level labels, thus eliminating the need for patch-level annotations. Typically, the MIL process involves tiling WSIs into patches, followed by processing these patches through a pre-trained feature extractor to generate either scalar outputs (instance-based MIL) or feature vectors (embedding-based MIL), and finally aggregating them for slide-level prediction. Among these approaches, embedding-based MIL, which provides a richer representation, has been found to be more effective in classifying WSIs \cite{Embedding_vs_Instance_MIL}.

However, training gigapixel WSIs using MIL poses significant memory constraints, often preventing end-to-end training. To mitigate this issue, feature extractor parameters are typically frozen to conserve GPU memory, which consequently limits the MIL aggregator's ability to identify key patches effectively. This problem is worsen by the reliance of state-of-the-art (SOTA) MIL models \cite{CLAM, TransMIL, DTFD-MIL} on the ResNet50 \cite{ResNet} supervised model, pre-trained on ImageNet-1K \cite{ImageNet-1K}, as the default feature extractor for benchmarking.

Advancements in self-supervised learning (SSL) have facilitated the development of pre-trained models tailored to histopathological representation using unlabeled datasets. In particular, DSMIL pre-trains a ResNet18 model, integrating multi-scale features into the MIL framework. TransPath \cite{TransPath} advances this approach by combining a CNN with Swin \cite{Swin} backbone, using semantically-relevant contrastive learning (SRCL) for better positive pair mining. Recent research \cite{Benchmarking_SSL_Histopathology} has also evaluated several SSL methods \cite{BarlowTwins,SwAV,MoCoV2,DINO} using ResNet50 and ViT-S/16 \cite{ViT} for patch classification and nuclei segmentation tasks.

\begin{figure*}[!htb]
\centering
\includegraphics[width=0.6\textwidth]{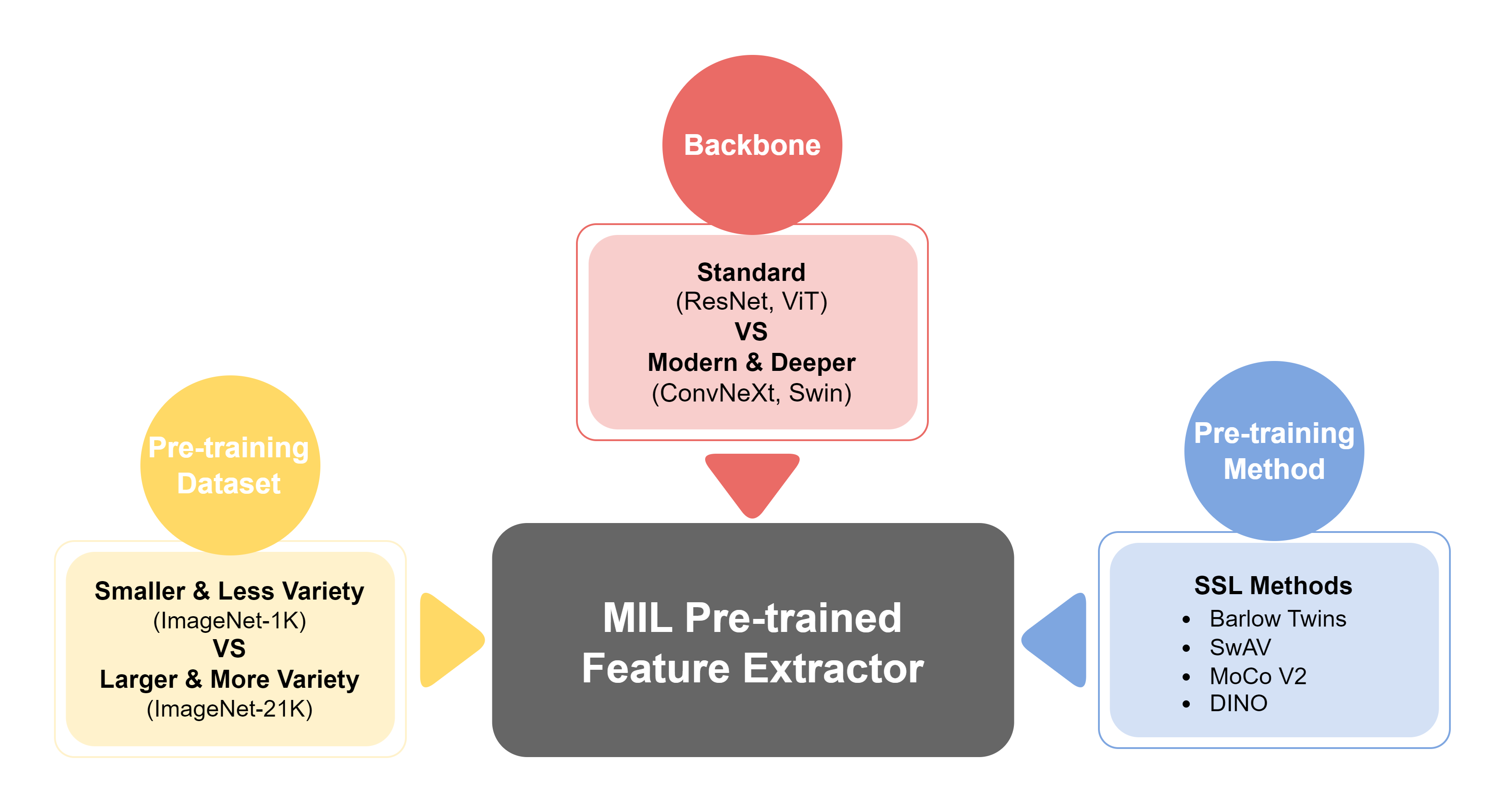}
\caption{Analytical setup overview for optimizing MIL pre-trained feature extractors.}
\label{fig:introduction}
\end{figure*}

Despite advances in MIL-based WSI classification, model performance often relies heavily on feature extractors that may not be optimized for the specific demands of histopathological data. Recent approaches have explored initializing it with foundation models \cite{TransPath, Benchmarking_SSL_Histopathology, UNI}, utilizing varied pre-training datasets (ImageNet-1K, in-domain), backbones (ResNet, ViT), and pre-training methods (supervised, SSL methods) to enhance WSI classification performance. \textit{Despite these variations, there is no clear guidance on selecting the optimal feature extractors to maximize WSI classification outcomes}. We argue that the existing studies may not fully leverage these strategies' benefits, suggesting that improved guidance could unlock greater potential in enhancing WSI classification performance.

Hence, the main contribution of this paper is a novel analysis of selecting optimal feature extractors and its impact on the performance of SOTA MIL models. Inspired by \cite{Broad_Study_Pretraining_DA_DG}, our analysis spans three dimensions: pre-training dataset, backbone, and pre-training method. Using two public WSI datasets, TCGA-NSCLC and Camelyon16 \cite{Camelyon16}, and employing four SOTA MIL models—ABMIL \cite{ABMIL}, DSMIL \cite{DSMIL}, TransMIL \cite{TransMIL}, and DTFD-MIL \cite{DTFD-MIL}—this study is the first to undertake such a comprehensive analysis focused on optimal feature extractor selection.

\begin{table*}[!htb]
\scriptsize
\centering
\caption{Performance comparison of SOTA MIL models when using modern and deeper backbones as feature extractors (ConvNeXt-B for CNNs and Swin-B for Transformer based models) against standard backbones (ResNet50 and ViT-S/16).}
\label{tab:standard_vs_modern_backbones_tcga_nsclc_camelyon16}
\begin{tabular}{|l|cc|cc|cc|cc|cc|cc|cc|cc|}
\hline
\multirow{3}{*}{\textbf{MIL Model}} & \multicolumn{8}{c|}{\textbf{TCGA-NSCLC}} & \multicolumn{8}{c|}{\textbf{Camelyon16}} \\
\cline{2-17}
& \multicolumn{2}{c|}{\textbf{ResNet50}} & \multicolumn{2}{c|}{\textbf{ConvNeXt-B}} & \multicolumn{2}{c|}{\textbf{ViT-S/16}} & \multicolumn{2}{c|}{\textbf{Swin-B}} & \multicolumn{2}{c|}{\textbf{ResNet50}} & \multicolumn{2}{c|}{\textbf{ConvNeXt-B}} & \multicolumn{2}{c|}{\textbf{ViT-S/16}} & \multicolumn{2}{c|}{\textbf{Swin-B}} \\
& \multicolumn{2}{c|}{\textbf{8.5M, \textcolor{blue}{IN-1K}}} & \multicolumn{2}{c|}{\textbf{43.2M, IN-21K}} & \multicolumn{2}{c|}{\textbf{21.7M, IN-21K}} & \multicolumn{2}{c|}{\textbf{86.7M, IN-21K}} & \multicolumn{2}{c|}{\textbf{8.5M, \textcolor{blue}{IN-1K}}} & \multicolumn{2}{c|}{\textbf{43.2M, IN-21K}} & \multicolumn{2}{c|}{\textbf{21.7M, IN-21K}} & \multicolumn{2}{c|}{\textbf{86.7M, IN-21K}} \\
\cline{2-17}
& ACC & AUC & ACC & AUC & ACC & AUC & ACC & AUC & ACC & AUC & ACC & AUC & ACC & AUC & ACC & AUC \\
\hline
ABMIL & 85.08 & 89.90 & \textbf{87.94} & \underline{92.97} & 84.60 & \textbf{93.13} & \underline{85.72} & 90.12 & \underline{78.30} & \underline{74.84} & 75.64 & 61.79 & 78.04 & 72.61 & \textbf{81.65} & \textbf{77.62} \\
DSMIL & \underline{84.29} & 91.24 & 84.28 & \underline{92.70} & 81.27 & 90.32 & \textbf{86.35} & \textbf{93.60} & \textbf{80.36} & \textbf{77.57} & \underline{79.33} & 75.01 & 73.64 & \underline{75.48} & 74.68 & 73.11 \\
TransMIL  & 85.08 & 90.96 & 88.09 & \underline{93.44} & \underline{88.10} & \textbf{93.95} & \textbf{89.37} & 89.18 & \underline{80.62} & \underline{80.77} & 78.55 & 78.86 & 79.59 & 78.58 & \textbf{88.11} & \textbf{89.86} \\
DTFD-MIL & 87.78 & \underline{94.34} & \underline{88.41} & 94.12 & 87.14 & 93.07 & \textbf{89.21} & \textbf{94.52} & \underline{82.17} & \underline{86.53} & 80.28 & 85.12 & 80.62 & 81.89 & \textbf{88.11} & \textbf{90.47} \\
\hline
\textbf{Average} & 85.56 & 91.61 & \underline{87.18} & \textbf{93.31} & 85.28 & 92.62 & \textbf{87.66} & \underline{91.85} & \underline{80.36} & \underline{79.93} & \textbf{\textcolor{red}{78.45}} & \textbf{\textcolor{red}{75.20}} & 77.97 & 77.14 & \textbf{83.14} & \textbf{82.77} \\
\hline
\end{tabular}
\end{table*}

\section{Analysis Setup}
\label{sec:analysis_setup}

In the context of the embedding-based MIL approach, feature extraction from the pre-trained model $f_p$ (mostly ResNet \cite{ResNet} and ViT \cite{ViT}) results in $F_i \in \mathbb{R}^{n(i) \times d}$, where $n(i)$ corresponds to the number of patches in WSI $i$, and $d$ varies with the chosen backbones. This study analyzes the comparative efficacy of diverse pre-trained feature extractors in WSI classification with its focus on the analysis of pre-training datasets, backbone designs, and pre-training methods (refer to Fig. \ref{fig:introduction} for details).

\textbf{Pre-training Dataset.} Most SOTA MIL models \cite{CLAM,TransMIL,DTFD-MIL} typically utilize feature extractors pre-trained on the ImageNet-1K dataset \cite{ImageNet-1K}, which comprises 1.2 million images across 1,000 classes and is commonly used for pre-training. In \cite{Swin,ConvNeXt}, exploring pre-training on ImageNet-21K \cite{ImageNet-21K}, a more extensive collection from ImageNet that includes 14,197,122 images divided into 21,841 classes, has shown improved transferability and generally better performance on the downstream ImageNet-1K classification. Given this improvement shown in the general context, we examine the untapped potential of a positive correlation between downstream WSI classification performance and the use of larger and more varied pre-training datasets (e.g., ImageNet-1K vs. ImageNet-21K) in enhancing effectiveness for WSI classification. This comparative analysis is conducted under the same backbones for both CNNs and Transformer-based.

\textbf{Backbone.} We investigate the effects of incorporating modern, larger, and more powerful backbones, moving beyond standard choices such as the CNN-based ResNet and Transformer-based ViT. Our goal is to examine whether these advanced backbones, when pre-trained with the same dataset and method, can generate improved features that lead to MIL models that are more robust and generalize better. For the CNN-based representative, we select ConvNeXt-B \cite{ConvNeXt}, which enhances traditional CNNs to better capture global dependencies by incorporating design elements from the ViT model. For the Transformer-based backbone, we opt for Swin-B \cite{Swin}, recognized for its incorporation of local context and efficient scalability.

\textbf{Pre-training Method.} Designing pre-trained models on in-domain datasets has become prevalent, with the work that showed pathology data pre-training surpasses ImageNet pre training \cite{Benchmarking_SSL_Histopathology}. Unlike supervised learning, SSL effectively leverages vast amounts of unlabeled data and offers clear advantages particularly in medical settings, where labeled data are scarce and require expert knowledge. Moreover, SSL is recognized for its ability to learn rich feature representations and enhancing generalization to new datasets. However, given the variety of available SSL methods, it remains unclear which method is the most suited for pre-trained feature extractors to enhance the performance of MIL models. Consequently, we investigate the impact of four representative SSL approaches: contrastive learning (MoCoV2 \cite{MoCoV2}), non-contrastive learning (Barlow Twins \cite{BarlowTwins}), clustering (SwAV \cite{SwAV}), and SSL with ViT (DINO \cite{DINO}).


\section{Experiments and Discussion}
\label{sec:experiments_discussion}

\subsection{Datasets}

We employ two publicly available WSI datasets: TCGA-NSCLC and Camelyon16 \cite{Camelyon16}. \textbf{TCGA-NSCLC} includes 1,046 WSIs (train: 836, test: 210), comprising 534 Lung Adenocarcinoma (LUAD) and 512 Lung Squamous Cell Carcinoma (LUSC) slides. Following \cite{DSMIL}, this dataset contains approximately 5.2 million $224 \times 224$ pixel patches, with around 5,000 patches per slide. \textbf{Camelyon16} is designed for detecting cancer metastases in lymph nodes ($<10\%$ per slide) and includes 399 WSIs (train: 270, test: 129) with imbalance training and test sets. To standardize tissue patch extraction and address magnification variations, we employ HS2P (based on CLAM \cite{CLAM}). With a pixel spacing set to 0.5 µm/px (20x magnification) and a tissue threshold of 0.1, we obtain about 4.8 million non-overlapping $256 \times 256$ pixel patches, averaging 12,000 patches per slide.

\subsection{Implementation Details}

Before feature extraction, patches are resized to $224 \times 224$ pixels and normalized using ImageNet-1K standards \cite{ImageNet-1K}, while pathology foundation models follow their specific normalization standards \cite{Benchmarking_SSL_Histopathology}. Feature vectors are obtained by removing the final fully connected layers or heads and applying global average pooling as needed. In ResNet50 models \cite{ResNet}, we follow the common practice \cite{CLAM,TransMIL,DTFD-MIL} of excluding the last convolutional module, reducing feature vector dimensionality from 2048 to 1024. The feature dimensionality ranges from 384 to 1024, depending on the backbone used. Using these extracted features, we train four SOTA MIL models: ABMIL \cite{ABMIL}, DSMIL \cite{DSMIL}, TransMIL \cite{TransMIL}, and DTFD-MIL (MaxMinS) \cite{DTFD-MIL}. Each model is trained for 50 epochs with a batch size of 1. All models except TransMIL use the Adam optimizer with a learning rate and weight decay of $1e-4$, while TransMIL uses the Lookahead optimizer with a learning rate of $2e-4$ and weight decay of $1e-5$. Performance is evaluated using accuracy (ACC) and area under the curve (AUC) metrics, with the mean calculated over \textbf{three runs} to mitigate initialization randomness. The experiments are conducted using PyTorch on a Linux Ubuntu 20.04.6 server equipped with two NVIDIA A100 GPUs (40GB each).

\begin{table*}[!htb]
\scriptsize
\centering
\caption{Effectiveness of SSL pre-training methods using the same in-domain pre-training datasets.}
\label{tab:ssl_pretraining_methods_tcga_nsclc_camelyon16}
\begin{tabular}{|l|cc|cc|cc|cc|cc|cc|cc|cc|}
\hline
\multirow{4}{*}{\textbf{MIL Model}} & \multicolumn{8}{c|}{\textbf{TCGA-NSCLC}} & \multicolumn{8}{c|}{\textbf{Camelyon16}} \\
\cline{2-17}
& \multicolumn{2}{c|}{\textbf{Barlow Twins}} & \multicolumn{2}{c|}{\textbf{SwAV}} & \multicolumn{2}{c|}{\textbf{MoCo V2}} & \multicolumn{2}{c|}{\textbf{DINO}} & \multicolumn{2}{c|}{\textbf{Barlow Twins}} & \multicolumn{2}{c|}{\textbf{SwAV}} & \multicolumn{2}{c|}{\textbf{MoCo V2}} & \multicolumn{2}{c|}{\textbf{DINO}} \\
& \multicolumn{2}{c|}{\textbf{ResNet50}} & \multicolumn{2}{c|}{\textbf{ResNet50}} & \multicolumn{2}{c|}{\textbf{ResNet50}} & \multicolumn{2}{c|}{\textbf{ViT-S/16}} & \multicolumn{2}{c|}{\textbf{ResNet50}} & \multicolumn{2}{c|}{\textbf{ResNet50}} & \multicolumn{2}{c|}{\textbf{ResNet50}} & \multicolumn{2}{c|}{\textbf{ViT-S/16}} \\
\cline{2-17}
& ACC & AUC & ACC & AUC & ACC & AUC & ACC & AUC & ACC & AUC & ACC & AUC & ACC & AUC & ACC & AUC \\
\hline
ABMIL & \underline{87.78} & \underline{93.99} & 85.87 & 93.27 & 85.71 & 89.86 & \textbf{90.48} & \textbf{96.83} & 91.47 & 92.06 & \textbf{94.83} & \textbf{95.14} & 76.74 & 73.03 & \underline{94.06} & \underline{94.57} \\
DSMIL & \underline{86.35} & \underline{93.97} & 85.72 & 93.53 & 76.03 & 89.08 & \textbf{89.05} & \textbf{96.34} & 88.63 & 87.88 & \underline{92.25} & \underline{91.26} & 65.38 & 66.95 & \textbf{95.09} & \textbf{98.25} \\
TransMIL & 89.68 & 92.11 & \underline{89.84} & \textbf{95.69} & 88.41 & 92.55 & \textbf{92.86} & \underline{95.64} & 93.28 & 94.26 & \underline{94.83} & \underline{96.64} & 93.02 & 94.4 & \textbf{97.15} & \textbf{98.10} \\
DTFD-MIL & 89.21 & 91.97 & \underline{89.52} & \underline{95.66} & 70.32 & 76.82 & \textbf{93.18} & \textbf{97.62} & 91.18 & 94.97 & \underline{94.83} & \underline{96.46} & 64.60 & 63.16 & \textbf{97.41} & \textbf{98.07} \\
\hline
\textbf{Average} & \underline{88.26} & 93.01 & 87.74 & \underline{94.54} & \textbf{\textcolor{red}{80.12}} & \textbf{\textcolor{red}{87.08}} & \textbf{91.39} & \textbf{96.61} & 91.14 & 92.29 & \underline{94.19} & \underline{94.92} & \textbf{\textcolor{red}{74.94}} & \textbf{\textcolor{red}{74.37}} & \textbf{95.93} & \textbf{97.25} \\
\hline
\end{tabular}
\end{table*}

\begin{figure}[!htb]
\centering
\includegraphics[width=8.5cm]{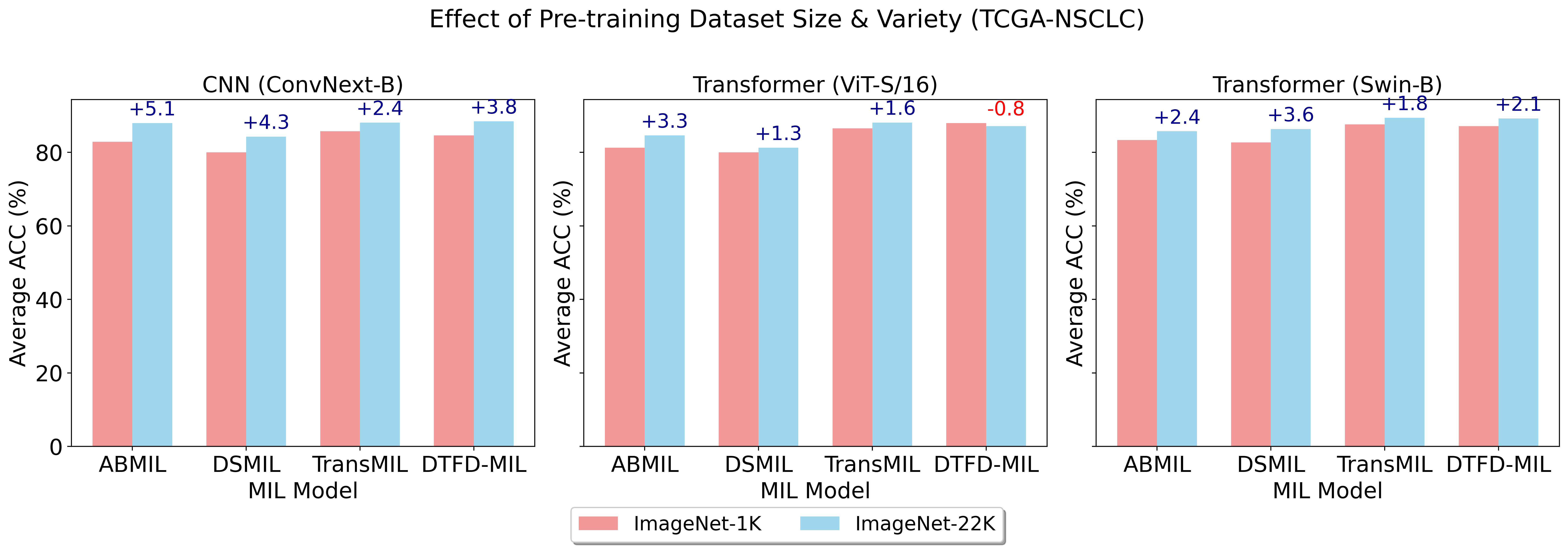}
\caption{The effect of using a larger and more varied pre-training dataset (ImageNet-1K \cite{ImageNet-1K} vs. ImageNet-21K \cite{ImageNet-21K}) on CNN and Transformer backbones of feature extractors.}
\label{fig:pretraining_dataset_size_and_variety}
\end{figure}

\subsection{Pre-training Dataset Size and Variety}

We investigate how the size and variety of a larger pre-training dataset for the feature extractors affect the performance of SOTA MIL models \cite{ABMIL,DSMIL,TransMIL,DTFD-MIL}. Comparing models pre-trained on ImageNet-1K \cite{ImageNet-1K} with those on ImageNet-21K \cite{ImageNet-21K}, we observe consistent performance improvements in WSI classification when ImageNet-21K is used for pre-training across various backbones for both CNN-based (ConvNeXt-B \cite{ConvNeXt}) and Transformer-based backbones (ViT-S/16 \cite{ViT}, Swin-B \cite{Swin}), 
as shown in Fig. \ref{fig:pretraining_dataset_size_and_variety}, with the exception of a small decrease for the ViT-S/16 backbone in DTFD-MIL. The performance boost is backbone-independent, suggesting a universal benefit from larger pre-training datasets, which offer richer features and stronger generalization capabilities.

\subsection{Standard vs Modern Backbones}
\label{subsec:standard_vs_modern_backbones}

Table \ref{tab:standard_vs_modern_backbones_tcga_nsclc_camelyon16} shows that modern and deeper backbones, such as ConvNeXt-B and Swin-B, improve performance over standard backbones (ResNet50, ViT-S/16) on the TCGA-NSCLC dataset. However, on the more challenging Camelyon16 dataset \cite{Camelyon16}, which includes small tumor regions, ConvNeXt-B underperforms relative to ResNet50, while Swin-B consistently outperforms both ResNet50 and ViT-S/16. This disparity may be attributed to the architectural strengths of Transformer-based models like Swin-B, which excel in handling complex spatial relationships within small ROIs through self-attention mechanisms that capture long-range dependencies. In contrast, ConvNeXt-B, despite its advanced CNN architecture, may lack the same level of adaptability for modeling fine-grained spatial patterns. These results suggest that Transformer-based models offer more reliable improvements, especially when scaling model parameters, due to their adaptive receptive fields and robust context-aware representations essential for distinguishing subtle morphological differences.

\subsection{Self-Supervised Pre-training Methods}

We evaluate the effectiveness of various SSL methods using the same pre-training dataset employed in pathology foundation models \cite{Benchmarking_SSL_Histopathology} on the TCGA-NSCLC and Camelyon16 datasets. As shown in Table \ref{tab:ssl_pretraining_methods_tcga_nsclc_camelyon16}, the choice of SSL method and backbone significantly influences WSI classification performance, with DINO on ViT-S/16 consistently outperforming contrastive learning (MoCoV2 \cite{MoCoV2}), non-contrastive (Barlow Twins \cite{BarlowTwins}), and clustering (SwAV \cite{SwAV}) approaches, while MoCoV2 on ResNet50 performs the least favorably. These results support previous findings on the superior capacity of Transformer-based models over CNNs for WSI tasks (see Section \ref{subsec:standard_vs_modern_backbones}) and suggest that contrastive pre-training methods like MoCoV2 are not effective for WSI classification. MoCoV2’s contrastive objective—bringing positive pairs closer while pushing apart negative pairs—struggles with high tissue similarity and class imbalance, which can degrade performance, particularly on challenging datasets like Camelyon16 with small ROIs. Furthermore, our analysis shows that selecting an appropriate pre-training method is more crucial than simply using an in-domain pre-training dataset. Specifically, an ineffective choice of pre-training method (MoCoV2) can yield worse results on in-domain data compared to models pre-trained on ImageNet (see Table \ref{tab:standard_vs_modern_backbones_tcga_nsclc_camelyon16} for comparison).

\section{Conclusion}
\label{sec:conclusion}

In this study, we have investigated feature extractor selection and design to enhance WSI classification within the MIL framework. Our evaluation across pre-training datasets, backbone models, and pre-training methods demonstrates that a carefully chosen SSL method has a greater impact on performance than an in-domain pre-training dataset alone. We recommend prioritizing Transformer-based backbones with deeper architectures over CNNs and using larger, more diverse pre-training datasets. These findings offer practical guidance for feature extractor selection and may help explain the superior performance of current SOTA pathology foundation models \cite{TransPath, UNI}, which leverage deeper Transformer backbones (e.g., Swin Tiny, ViT Large), advanced SSL techniques beyond basic contrastive learning, and diverse cancer-type pre-training datasets. We hope these insights will support the development of more effective pathology foundation models.

\section{Compliance with Ethical Standards}

This research study was conducted retrospectively using human subject data made available in open access by The Cancer Genome Atlas Program (TCGA) and \cite{Camelyon16}. Ethical approval was not required as confirmed by the license attached with the open access data.

\section{Acknowledgements}

This research was supported by the Seegene Medical Foundation, South Korea, under the project “Development of a Multimodal Artificial Intelligence-Based Computer-Aided Diagnosis System for Gastrointestinal Endoscopic Biopsies” (Grant Number: G01240151).

\bibliographystyle{IEEEbib}
\bibliography{strings,refs}

\end{document}